\documentclass[a4paper]{article}

\usepackage[dvipdfmx]{graphicx}
\usepackage{multirow}
\usepackage{lscape}
\usepackage{siunitx}
\usepackage{amsmath}
\usepackage{wrapfig}
\usepackage[misc]{ifsym}
\usepackage[dvipdfmx]{graphicx}
\usepackage{mathtools}
\usepackage{amsmath,amssymb}
\usepackage{multirow}
\usepackage{xr}
\usepackage[hang,small,bf]{caption}
\usepackage[subrefformat=parens]{subcaption}
\usepackage{authblk}

\DeclareMathOperator{\round}{ROUND}
\DeclareMathOperator{\qscale}{\Delta}
\DeclareMathOperator{\maxvalue}{a}
\DeclareMathOperator{\minvalue}{b}

\def\vect#1{\mbox{\boldmath $#1$}}

\usepackage{multirow}
\usepackage{amsmath}
\usepackage{capt-of}
\usepackage{tabularx}
\usepackage{epsfig}
\usepackage{amssymb}
\usepackage{amsfonts}
\usepackage{booktabs}
\usepackage{scalerel}
\usepackage[inline]{enumitem}
\usepackage{listings}
\usepackage{varwidth}
\usepackage[export]{adjustbox}
\usepackage{tikz}
\usetikzlibrary{tikzmark}

\usepackage{stmaryrd}
\usepackage{bbm}
\usepackage{wrapfig}
\usepackage{pifont}

\definecolor{deepblue}{rgb}{0,0,0.5}
\definecolor{officeblue}{RGB}{0,102,204}
\definecolor{deepred}{rgb}{0.6,0,0}
\definecolor{deepgreen}{rgb}{0,0.5,0}
\definecolor{mybrickred}{RGB}{182,50,28}

\definecolor{fillcolor}{RGB}{216,217,252}

\usepackage{etoolbox}
\usepackage{framed}

\newif\ifxetexorluatex
\ifxetex
  \xetexorluatextrue
\else
  \ifluatex
    \xetexorluatextrue
  \else
    \xetexorluatexfalse
  \fi
\fi
\newcommand*\quotesize{60} 
\newcommand*{\openquote}
   {\tikz[remember picture,overlay,xshift=-4ex,yshift=-2.5ex]
   \node (OQ) {\fontsize{\quotesize}{\quotesize}\selectfont``};\kern0pt}

\newcommand*{\closequote}[1]
  {\tikz[remember picture,overlay,xshift=4ex,yshift={#1}]
   \node (CQ) {\fontsize{\quotesize}{\quotesize}\selectfont''};}

\colorlet{shadecolor}{white}

\newcommand*\shadedauthorformat{\emph} 

\newcommand*\authoralign[1]{%
  \if#1l
    \def\authorfill{}\def\quotefill{\hfill}
  \else
    \if#1r
      \def\authorfill{\hfill}\def\quotefill{}
    \else
      \if#1c
        \gdef\authorfill{\hfill}\def\quotefill{\hfill}
      \else\typeout{Invalid option}
      \fi
    \fi
  \fi}
\newenvironment{shadequote}[2][l]%
{\authoralign{#1}
\ifblank{#2}
   {\def\shadequoteauthor{}\def\yshift{-2ex}\def\quotefill{\hfill}}
   {\def\shadequoteauthor{\par\authorfill\shadedauthorformat{#2}}\def\yshift{2ex}}
\begin{snugshade}\begin{quote}\openquote}
{\shadequoteauthor\quotefill\closequote{\yshift}\end{quote}\end{snugshade}}

\title{\bf Magic for the Age of Quantized DNNs}
\date{}

\author[1]{Yoshihide Sawada\thanks{Equal contribution.\\ Preprint. Work in progress.}}
\author[$\ast$1,2]{Ryuji Saiin}
\author[2]{Kazuma Suetake}

\affil[1]{Tokyo Research Center, AISIN, Tokyo, Japan}
\affil[2]{AISIN SOFTWARE, Aichi, Japan}

\begin{document}

\maketitle

\begin{abstract}
Recently, the number of parameters in DNNs has explosively increased, as exemplified by LLMs~(Large Language Models), making inference on small-scale computers more difficult.
Model compression technology is, therefore, essential for integration into products.
In this paper, we propose a method of quantization-aware training.
We introduce a novel normalization~(Layer-Batch Normalization) that is independent of the mini-batch size and does not require any additional computation cost during inference. 
Then, we quantize the weights by the scaled round-clip function with the weight standardization.
We also quantize activation functions using the same function and apply surrogate gradients to train the model with both quantized weights and the quantized activation functions.
We call this method {\it Magic for the age of Quantised DNNs~(MaQD)}. 
Experimental results show that our quantization method can be achieved with minimal accuracy degradation.
\end{abstract}
\section{Introduction} 
\label{sec:introduction}
\begin{shadequote}[r]{\small Arthur C. Clarke~\cite{clarke1962hazards}}
{\small \it Any sufficiently advanced technology is indistinguishable from magic.}
\end{shadequote}
As computational power increases, the trend toward training large-scale deep neural networks~(DNNs) on large-scale datasets continues unabated~\cite{brown2020language,kirillov2023segment,team2023gemini}, and the use of cloud computing is essential for such large-scale DNNs. 
However, when real-time processing is required, such as in automatic driving peripheral recognition, online processing becomes difficult with the use of cloud computing.
This implies that technology is required to integrate large-scale DNNs into small IoT devices, such as those based on microcontroller units~\cite{lin2021memory,lin2020mcunet,lin2022device}.
Namely, model compression techniques, as in \cite{hubara2018quantized,jacob2018quantization,ma2024era,wang2023bitnet}, for integration into hardware will become more critical.

\begin{figure}[t]
  \begin{minipage}[t]{0.3\linewidth}
    \centering
    \includegraphics[keepaspectratio, scale=0.32]{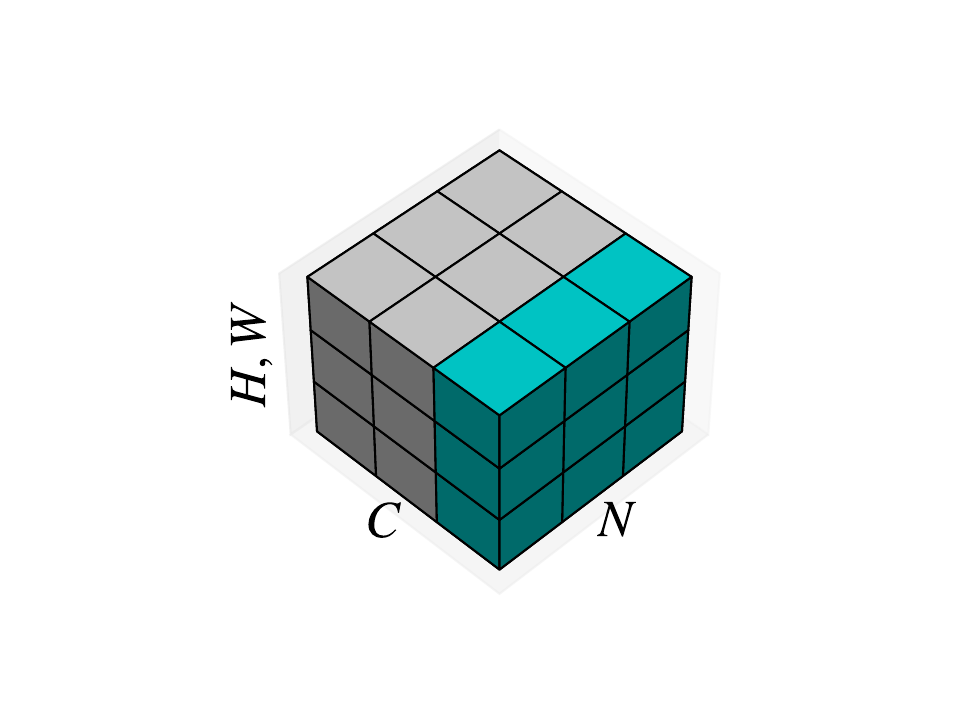}
  \end{minipage}
  \begin{minipage}[t]{0.3\linewidth}
    \centering
    \includegraphics[keepaspectratio, scale=0.32]{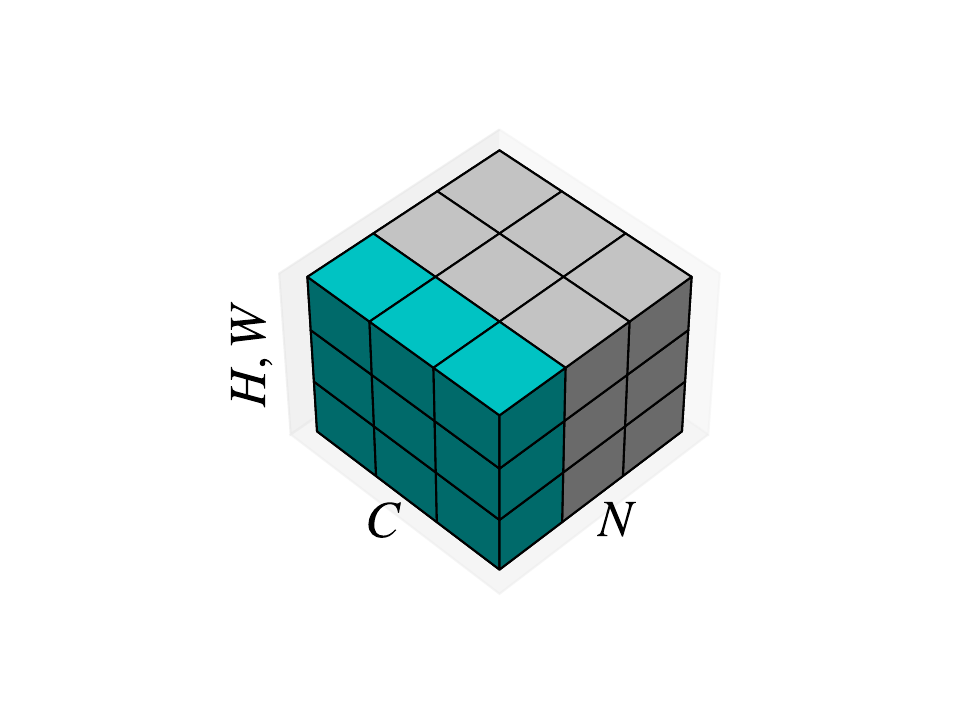}
  \end{minipage}
  \begin{minipage}[t]{0.3\linewidth}
    \centering
    \includegraphics[keepaspectratio, scale=0.32]{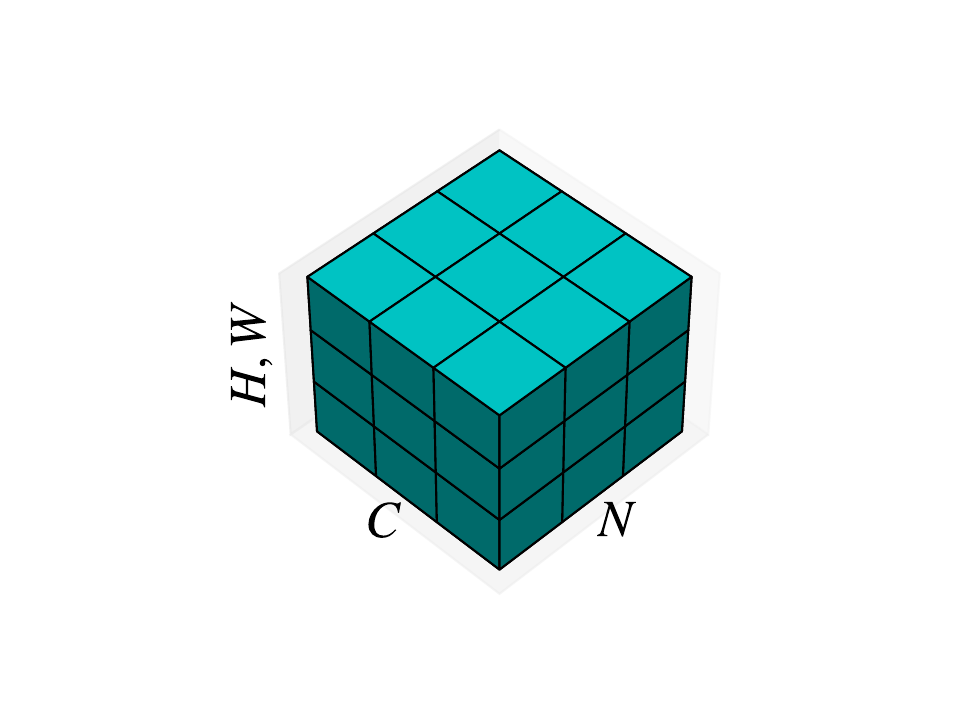}
  \end{minipage}
  \vspace{-7mm}
  \caption{Range per normalization layer. Green areas are used for normalization, where $C$ denotes the channel axis, $N$ denotes the batch axes, and $H$ and $W$ denote the height and width axies, respectively. From left to right, BN, LN, and LBN are represented, respectively.}
  \label{fig:lbn}
\end{figure}

We are considering ways to quantize DNNs while training rather than quantizing the trained model, as we expect that the quantization-aware training has a better trade-off between compression efficiency and inference accuracy.
To achieve efficient quantization, we introduce a novel normalization technique called Layer-Batch Normalization~(LBN), which is independent of the mini-batch size, unlike Batch Normalization~(BN)~\cite{ioffe2015batch}. 
LBN does not require computationally expensive expected value processes during inference for the normalization, unlike layer normalization~(LN)~\cite{ba2016layer}.
In addition, because the proposed method is independent of the mini-batch size, we can train DNNs with smaller mini-batch sizes while maintaining accuracy.
While BN recommends the use of large mini-batch sizes for training on large-scale datasets~\cite{smith2018don}, which has resulted in an increased demand for machine resources, the proposed method can help alleviate this resource requirement.
Figure~\ref{fig:lbn} represents the overview of the LBN.
On the basis of this normalization, weight standardization~(WS)~\cite{qiao2019micro} and quantization by scaled round-clip functions and surrogate gradients~\cite{bengio2013estimating,courbariaux2016binarized,hubara2018quantized,suetake2023s3nn} can be combined to obtain the quantized DNNs with minimal accuracy degradation.
We call this method {\it Magic for the age of Quantized DNNs~(MaQD)}. 

Our contributions of this study include the following:
\renewcommand{\labelenumi}{(\arabic{enumi})}
\begin{enumerate}
    \item We propose a novel normalization, LBN, which is independent of the mini-batch size and does not require computationally expensive expected value processes during inference.
    \item We propose the MaQD, which combines the LBN, WS, scaled round-clip functions, and surrogate gradients, to train the quantized DNNs.
    \item Experimental results show that our quantization technique achieves a good trade-off between compression efficiency and inference accuracy.
\end{enumerate}

\section{Preliminaries}
In this section, we explain the traditional normalization~\cite{ba2016layer,ioffe2015batch} and WS~\cite{qiao2019micro}.

\subsection{Normalization Layer}
\label{sub:norm}
Let $a_{i,j,k}^{(l)} \in \mathbb{R}$ be the feature after weight transformation in the neural network, where $l \in \mathbb{N}$ represents the layer, $i \in \mathbb{N} \ (1 \leq i \leq C^{(l)})$ represents the channel, $(j, k) \in \mathbb{N}^2 \ (1 \leq j \leq H^{(l)}, 1 \leq k \leq W^{(l)})$ represents the two-dimensional coordinates~(because we consider the image classification tasks), and $C^{(l)}$, $W^{(l)}$, and $H^{(l)}$ represent the channel, width, and height of the feature map in $l$-th layer.
Then, the ordinary form of the normalization layer in the neural network, when $\mathbb{E}[\cdot]$ represents a certain expectation operator, is defined as follows:
\begin{align}
\begin{aligned}
\label{eq:normalization-general}
\hat{a}_{i,j,k} &:= \frac{g_{i,j,k}}{\sigma_{i,j,k}} (a_{i,j,k} - \mu_{i,j,k}) + b_{i,j,k}, \\
\mu_{i,j,k} &:= \mathbb{E}[a_{i,j,k}], \\
\sigma_{i,j,k} &:= \sqrt{\mathbb{E}[(a_{i,j,k} - \mu_{i,j,k})^2] + \epsilon}, \\
\end{aligned}
\end{align}
where $g_{i,j,k}, b_{i,j,k} \in \mathbb{R}$ are the trainable parameters, and $\epsilon > 0$ is the constant to prevent divide-by-zero.
Note that the normalization layer is applied to the features after the weight transformation and before the activation function, and the symbol $(l)$ is removed for the simplicity.
In addition, note that values can be pre-merged into other parameters~(e.g., weights) if the $\mu_{i,j,k}$ and $\sigma_{i,j,k}$ are independent of input data during inference (i.e., in the case when the values can be fixed during inference).

\subsubsection{Batch Normalization~(BN) and Layer Normalization~(LN)}
\label{subsub:bn_ln}
In this section, we explain the LN~\cite{ba2016layer} and BN~\cite{ioffe2015batch}, which are utilized by many DNNs, including Transformer~\cite{vaswani2017attention}, ResNet~\cite{he2016deep}, and DenseNet~\cite{huang2017densely}.
The following equations represent the BN and LN, respectively.
\begin{align}
\begin{aligned}
\label{eq:normalization-batch}
\mathbb{E}_{\rm BN}[z_{i,j,k}] &:= \frac{1}{HWN} \sum_{j=1}^{H} \sum_{k=1}^{W} \sum_{n=1}^{N} [z_{i,j,k}], \\
\end{aligned}
\end{align}
\begin{align}
\begin{aligned}
\label{eq:normalization-layer}
\mathbb{E}_{\rm LN}[z_{i,j,k}] &:= \frac{1}{CHW} \sum_{i=1}^{C} \sum_{j=1}^{H} \sum_{k=1}^{W} [z_{i,j,k}], \\
\end{aligned}
\end{align}
where $z_{i,j,k} = z_{i,j,k}(\vect{x}_n)$ represents a function depending on a sample $\vect{x}_n \ (1 \leq n \leq N)$ in the mini-batch.
For the trainable parameters $g_{i,j,k}$ and $b_{i,j,k}$, the BN provides only channel dependence, whereas the LN provides both channel and coordinate dependence.

\subsection{Weight Standardization~(WS)}
\label{sub:wn}
This section describes the standardization method for weights.
Let $W_{{\rm out}, {\rm in}} \in \mathbb{R}$ denote the weight parameter. Then, the WS is denoted as follows~\cite{qiao2019micro}:
\begin{align}
\begin{aligned}
\label{eq:weight-standardization}
\hat{W}_{{\rm out}, {\rm in}} &:= \frac{W_{{\rm out}, {\rm in}} - \mu_{W_{\rm out}}}{\sqrt{F_{\rm In}} \sigma_{W_{\rm out}} + \epsilon}, \\
\mu_{W_{\rm out}} &:= \frac{1}{F_{\rm In}} \sum_{{\rm in}=1}^{F_{\rm In}} W_{{\rm out}, {\rm in}}, \\
\sigma_{W_{\rm out}} &:= \sqrt{\frac{1}{F_{\rm In}} \sum_{{\rm in}=1}^{F_{\rm In}} (W_{{\rm out}, {\rm in}} - \mu_{W_{\rm out}})^2}, \\
\end{aligned}
\end{align}
where $F_{\rm In}$ represents the number of fan-in.

\section{Proposed Method}

\subsection{Layer-Batch Normalization~(LBN)}
\label{sub:lbn}
The BN is a standard method for obtaining stable and good convergence results when training DNNs.
However, to achieve this effect, ensuring a sufficiently large mini-batch size is necessary, which often requires a significant amount of memory capacity.
On the other hand, the LN does not cause mini-batch dependence but does not provide the same training efficiency as the BN.

To complement these normalization layers, we propose the LBN as follows:
\begin{align}
\begin{aligned}
\label{eq:normalization-layer-batch}
\mathbb{E}_{\rm LBN}[z_{i,j,k}] &:= \frac{1}{CHWN} \sum_{i=1}^{C} \sum_{j=1}^{H} \sum_{k=1}^{W} \sum_{n=1}^{N} [z_{i,j,k}]. \\
\end{aligned}
\end{align}
Note that, for the trainable parameters $g_{i,j,k}$ and $b_{i,j,k}$, only the channel dependence is given, as in the BN.

\subsection{Magic for the age of Quantized DNNs~(MaQD)}
\label{sub:maqd}

This section explains the quantization of weights and activation functions.
To quantize them, we use the following scaled round-clip function.
\begin{align}
\label{eq:general_quant}
Q(z, \qscale, \maxvalue, \minvalue) := \max \left(\maxvalue, \min \left(\minvalue, \frac{\round(\qscale z)}{\qscale}\right) \right),
\end{align}
where $z \in \mathbb{R}$ represents the input to this function, $\maxvalue, \minvalue \in \mathbb{R}$ represent the minimum and maximum values respectively, $\round(\cdot)$ represents the round function which outputs the integer nearest to the input, and $\qscale \in \mathbb{R}$ represents a value related to the number of states. 
Note that this function differs from \cite{ma2024era} in that it includes $\qscale$, which enables control over the degree of quantization.

Using Eq.~\ref{eq:general_quant}, qauantizations of the weights and activation functions are carried out as follows\footnote{We can select either $M_{\rm w} / 2$ or $(M_{\rm w} - 1)/ 2$.}.
\begin{align}
\label{eq:quant}
Q_{\rm w}(\hat{W}_{{\rm out}, {\rm in}}, M_{\rm w}) &:= Q\left(s\hat{W}_{{\rm out}, {\rm in}}, \frac{M_{\rm w}}{2} \ {\rm or} \ \frac{M_{\rm w} - 1}{2}, -1, 1 \right), \\
Q_{\rm a}(\hat{a}_{i,j,k}, M_{\rm a}) &:= Q\left(\hat{a}_{i,j,k}, M_{\rm a} - 1, 0, 1 \right),
\end{align}
where $M_{\rm w} \in 2 \mathbb{N} + 1$ and $M_{\rm a} \in \mathbb{N}_{\geq 2}$.
$s \in \mathbb{R}$ represents the scaling factor, and we set $s = 1/3$.
This value gives a confidence interval of 99.7\% if the distribution of weights corresponds to Gaussian.
$M_{\rm w}$ and $M_{\rm a}$ represent the number of states for weights and activation functions, respectively.
For example, if we quantize weights to 2 bits, we take values of $\{-1, 0, 1\}$ and set $M_w=3$. 
This setting corresponds to 1.58 bits in~\cite{ma2024era}.
Also, unlike quantization of weights, quantization of the activation function does not include values under zero.
Namely, if we set 2 bits, $Q_a(\hat{a}_{i,j,k}, M_a)$ takes values of $\{0, 1/3, 2/3, 1 \}$, and $M_a$ becomes $4 = 2^2$.
Note that we can achieve high computational efficiency and memory savings in such a quantized model by preparing additional floating scaling factors for each state during inference, as in~\cite{rastegari2016xnor}.

\subsection{Training Process}
\label{sub:training}
In this article, we focus on the image classification tasks using the loss function $L$~\cite{saiin2023spike,Xiao2022OnlineNetworks} described as follows:
\begin{align}
\begin{aligned}
L := (1-\gamma)\text{CE} + \gamma \text{MSE},
\end{aligned}
\end{align}
where $\text{CE}$ is the cross-entropy between the label and softmax output, and $\text{MSE}$ is the mean-squared error between the one-hot encoded label and logit of softmax output. 
$\gamma \in \mathbb{R}$ is the hyperparameter to balance two losses and we set $\gamma = 0.05$.
Note that while accuracy improvement through knowledge distillation~\cite{hinton2015distilling,okuno2021lossless,polino2018model} could easily be introduced to MaQD, we do not address this integration in this article as our focus is solely on the quantization itself.

\begin{figure}[t]
  \begin{minipage}[t]{0.5\linewidth}
    \centering
    \includegraphics[width=1.0\linewidth]{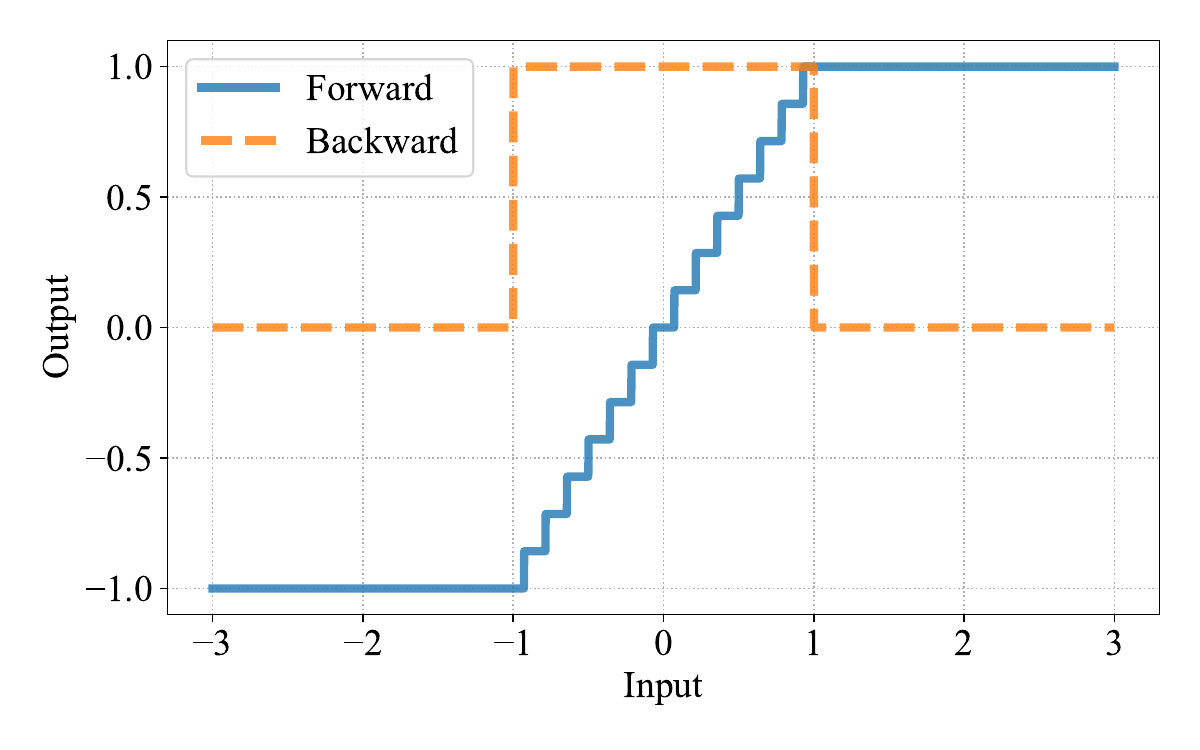}\\
    \vspace{-3mm}
    \subcaption{Weight}
  \end{minipage}
  \begin{minipage}[t]{0.5\linewidth}
    \centering
    \includegraphics[width=1.0\linewidth]{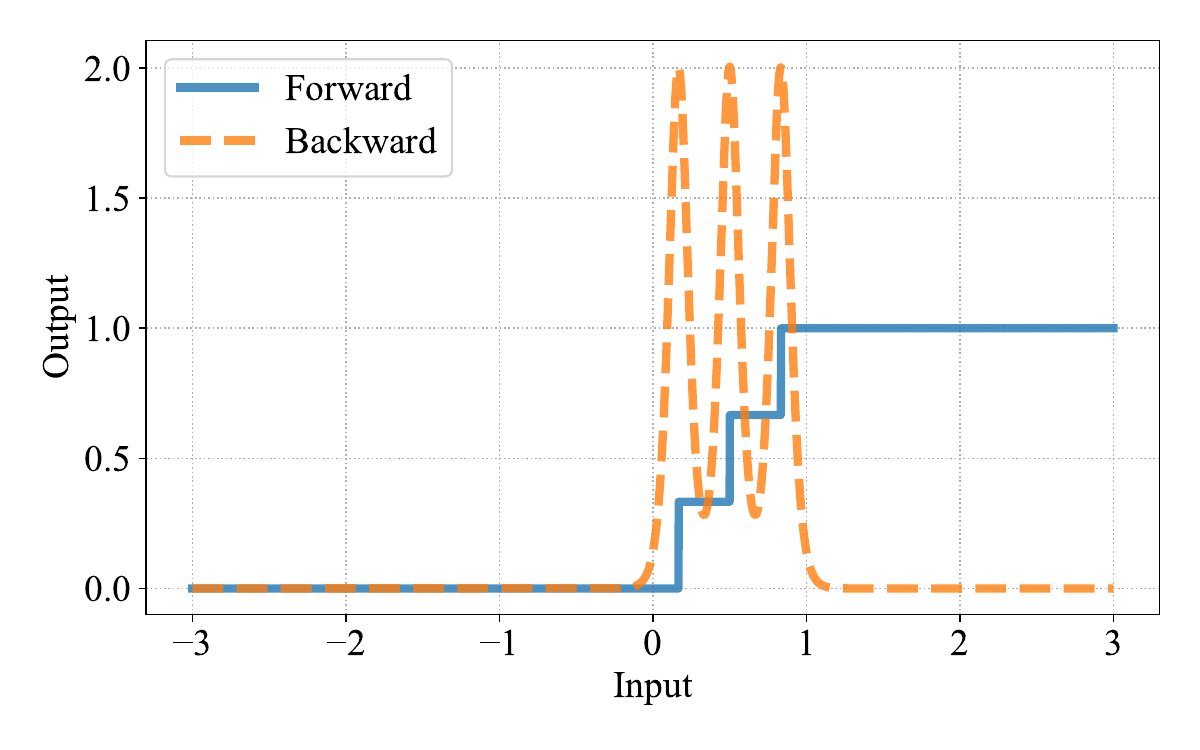}\\
    \vspace{-3mm}
    \subcaption{Activation function}
  \end{minipage}
  \caption{Surrogate gradients for weights and activation functions. The blue line represents forward propagation while the orange line represents backpropagation (surrogate gradient).}
  \label{fig:surrogate_gradients}
\end{figure}

To train models using the aforementioned loss function, differential calculations are necessary.
However, MaQD is clearly non-differentiable.
In this article, we adopt the surrogate gradient technique to approximate the gradient of the weights and activation functions as follows:
\begin{align}
\frac{\partial Q_{\rm w}(\hat{W}_{{\rm out}, {\rm in}}, M_{\rm w})}{\partial \hat{W}_{{\rm out}, {\rm in}}} &:= 1_{\left| s \hat{W}_{{\rm out}, {\rm in}} \right| < 1}(s \hat{W}_{{\rm out}, {\rm in}}), \\
\frac{\partial Q_{\rm a}(\hat{a}_{i,j,k}, M_{\rm a})}{\partial \hat{a}_{i,j,k}} &:= \sum_{m=1}^{M_{\rm a}-1} \frac{\partial \sigma_{\alpha}(\hat{a}_{i,j,k} - b_m)}{\partial \hat{a}_{i,j,k}},
\label{eqn:q_a}
\end{align}
where $1_X (\cdot)$ denotes the indicator function on the set $X$, and $\partial \sigma_{\alpha}(\hat{z}_{i,j,k} - b_m) / \partial \hat{z}_{i,j,k}$ is the gradient of the scaled sigmoid function~\cite{suetake2023s3nn}.
\begin{align}
\frac{\partial \sigma_{\alpha}(z)}{\partial z} &= \frac{1}{\alpha} \sigma_{\alpha}(z) (1 - \sigma_{\alpha}(z)), \\
\sigma_{\alpha}(z) &:= \frac{1}{1 + \exp{(-z/\alpha)}}.
\end{align}
Here, we set $\alpha = 0.25$.
$b_m \in \mathbb{R}$ in Eq.~\ref{eqn:q_a} represents the threshold changing from $m$-th state to $(m+1)$-th state~(e.g., the timing at which the output changes from 0 to 1 when we quantize to 1 bit), and the equation can be described as follows:
\begin{align}
b_m &= \frac{1}{\qscale}\left( \round_m(\qscale z) + \frac{1}{2} \right),\\
&= \frac{1}{M_{\rm a}-1}\left( (m-1) + \frac{1}{2} \right),
\end{align}
where $m \in \mathbb{N}_{< M_{\rm a}}$, and $\round_m(\cdot)$ denotes the $m$-th state~(i.e., $\round_m(\cdot) := m-1$).
Figure~\ref{fig:surrogate_gradients} shows each surrogate gradient.
Note that when the surrogate gradient of the activation function was set to Straight-Through Estimator~(STE)~\cite{bengio2013estimating,courbariaux2016binarized} as well as weights, the accuracy deteriorates significantly.

\section{Experiments}
\label{sec:experiment}
We first demonstrate the effectiveness of the LBN~(+WS) and then present the quantization results of the MaQD.

\subsection{Verification of the Efficacy of LBN}
\label{sub:superiority_lbn}
In this section, we compare the normalization techniques, i.e., LBN+WS, LBN, BN+WS, and BN.

\subsubsection{Settings}
\label{subsub:settings_lbn}
The following conditions were adopted to observe the differences between different normalizations when the mini-batch size varies:
\begin{itemize}
    \item Dataset: CIFAR-100~\cite{krizhevsky2009learning}
    \item Network architecture: Non-quantized 9-layer CNN
    \item Hyperparameters:
    \begin{itemize}
        \item Optimizer: Momentum SGD
        \item Scheduler: Cosine Annealing
        \item Learning rate, the number of epochs, and weight decay: The values obtained from the search with a mini-batch size of $128$ were used as baseline values, and for the learning rate, the baseline values were multiplied by $N/128$ depending on the mini-batch size $N$. In particular, the total number of updates was automatically adjusted as the mini-batch size changed, while keeping the number of epochs constant.
    \end{itemize}
\end{itemize}

\subsubsection{Results}
\label{subsub:reuslts_lbn}
\begin{figure}[t]
    \centering
    \includegraphics[width=0.45\linewidth]{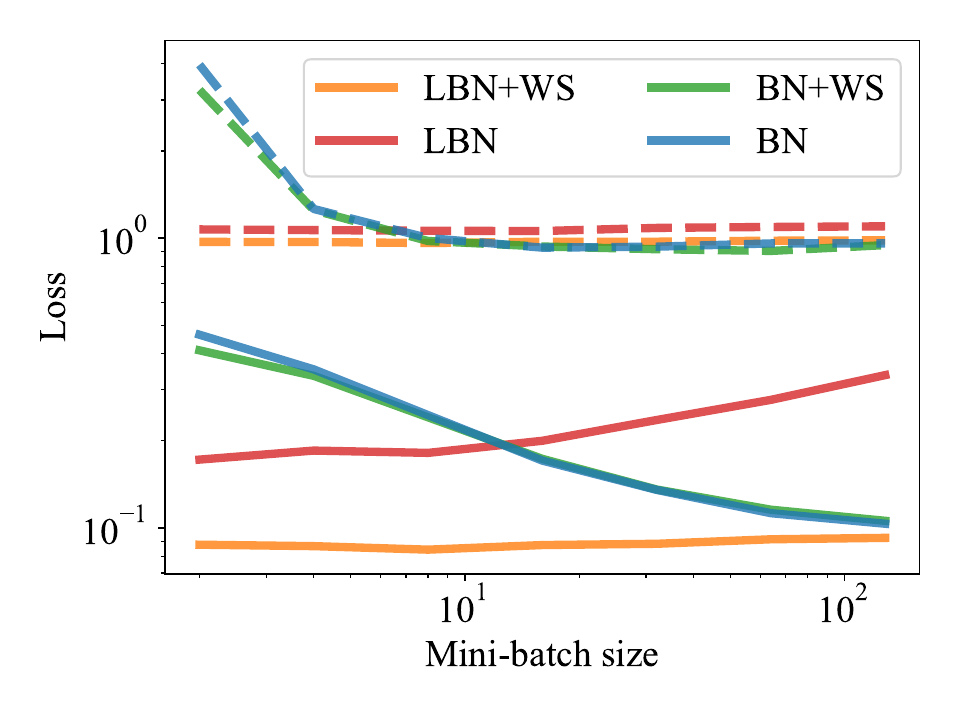}
    \includegraphics[width=0.45\linewidth]{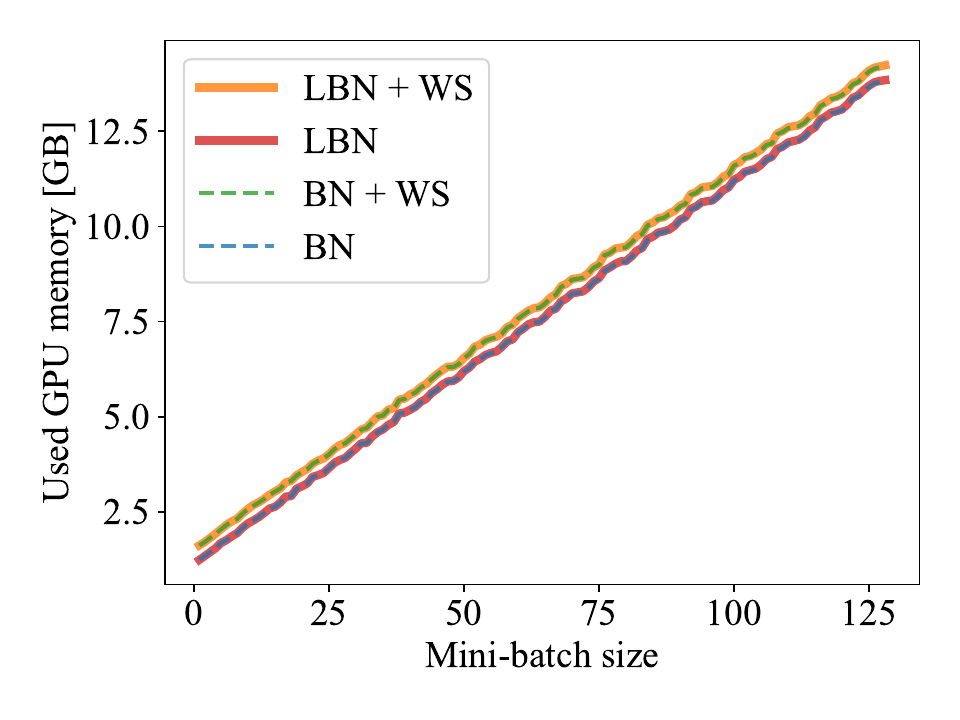}
    \vspace{-3mm}
    \caption{Left: Losses for training with varying mini-batch sizes for each normalization method. The solid and dashed lines represent the loss for the training and test datasets, respectively. Right: GPU memory usage when the mini-batch size is varied for each normalization method.}
    \label{fig:results_lbn}
\end{figure}

Figure~\ref{fig:results_lbn} shows the experimental results.
From the training results of each normalization method when varying the mini-batch size~(Fig.~\ref{fig:results_lbn}, left), the following can be observed:
\begin{itemize}
    \item LBN+WS achieves lower training losses than any of the other methods, regardless of the mini-batch size.
    \item LBN+WS exhibits less dependence on mini-batch size compared to BN, and loss difference between LBN+WS and BN diminishes as the mini-batch size increases.
\end{itemize}
Next, from the GPU memory usage during training for each normalization when varying the mini-batch size~(Fig.~\ref{fig:results_lbn}, right), the following can be observed:
\begin{itemize}
    \item GPU memory usage increases in proportion to the mini-batch size.
    \item GPU memory usage varies among the methods, with the order being \\
    BN $<$ LBN $<$ BN+WS $<$ LBN+WS 
    (although it may be difficult to discern from the figure, LBN uses slightly more GPU memory than BN).
\end{itemize}

These experimental results indicate that incorporating WS into BN does not enhance the independence of mini-batch size and that LBN without WS does not effectively reduce training and test losses.
Therefore, it is crucial to utilize both WS and LBN in conjunction.
Also, the impact on GPU memory usage is primarily determined by the mini-batch size rather than the choice of the normalization.
Therefore, since LBN is independent of the mini-batch size, we require less GPU memory than existing methods while achieving low training and test losses~(although LBN utilizes slightly more GPU memory than BN).

\subsection{Verification of the Efficacy of MaQD}
\label{sub:superiority_maqd}
As described above, the LBN+WS provides better results than BN, especially when we set a small mini-batch size. 
Thus, this section shows the quantization results using LBN+WS~(i.e., MaQD).

\subsubsection{Settings}
\label{subsub:settings_maqd}
The following conditions were adopted to observe the differences between different normalizations when varying the mini-batch size:
\begin{itemize}
    \item Dataset: CIFAR-10~\cite{krizhevsky2009learning} and CIFAR-100
    \item Network architecture: VGG and PreActResNet~(see Appendix for details)
    \item Hyperparameters:
    \begin{itemize}
        \item $M_{\rm w} / 2 \ {\rm or} \ (M_{\rm w} - 1)/ 2$ in Eq.~\ref{eq:quant}: $M_{\rm w} / 2$
        \item Optimizer: Momentum SGD
        \item Scheduler: Cosine Annealing
        \item Learning rate: $10^{-2}$
        \item The number of epochs: 300
        \item Mini-batch size: 100
    \end{itemize}
\end{itemize}

\subsubsection{Results}
\label{subsub:results_maqd}
\begin{figure}[t]
    \centering
    \includegraphics[width=0.45\linewidth]{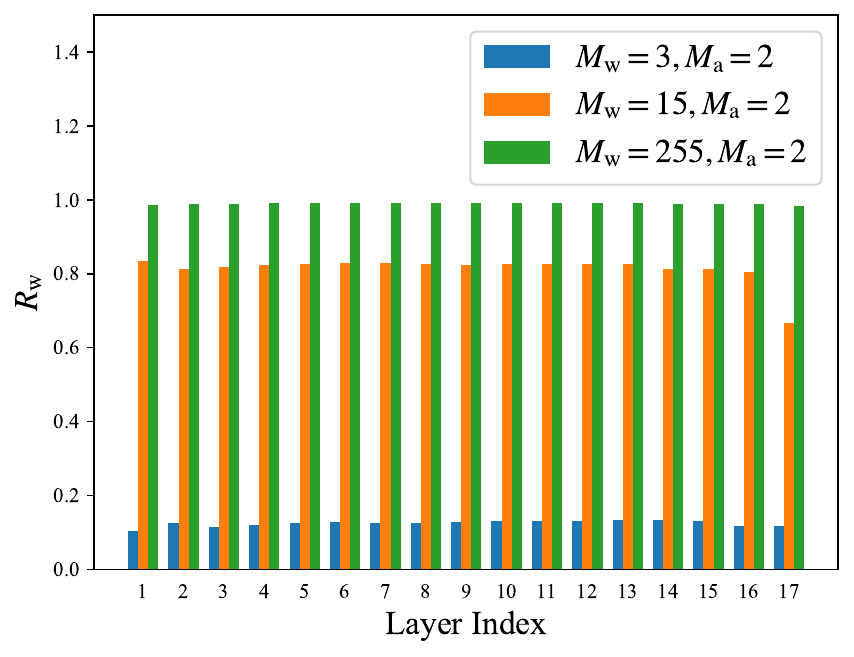}
    \includegraphics[width=0.45\linewidth]{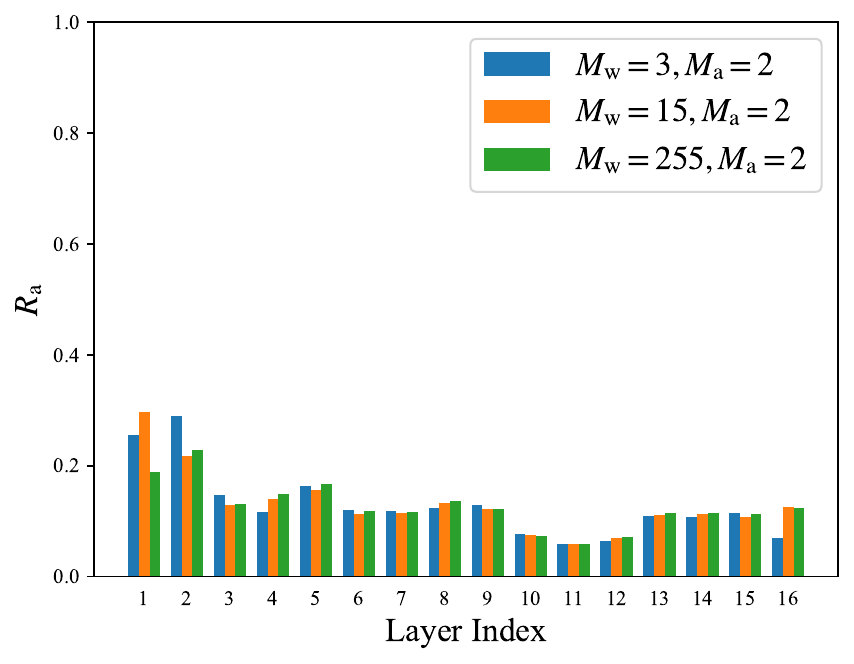}\\
    \includegraphics[width=0.45\linewidth]{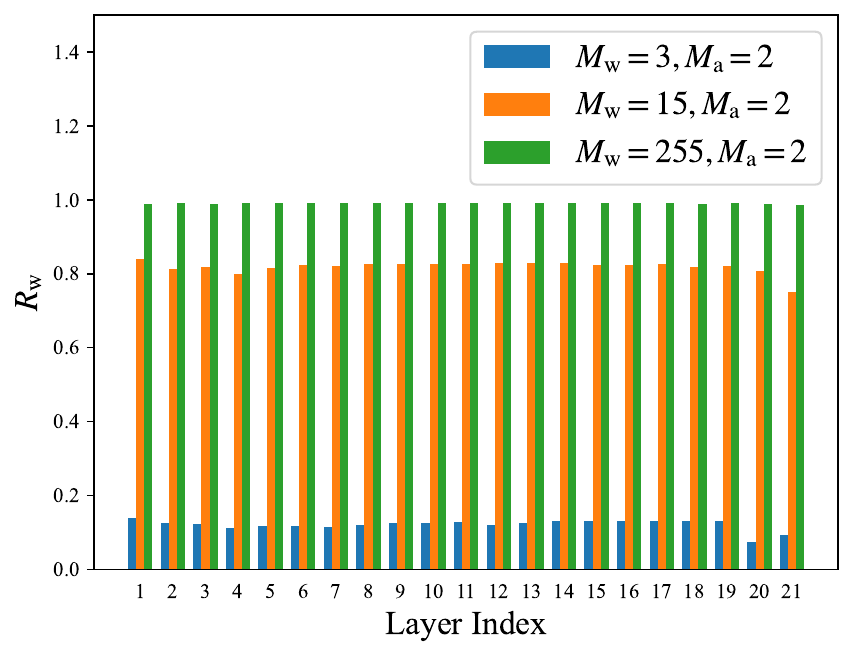}
    \includegraphics[width=0.45\linewidth]{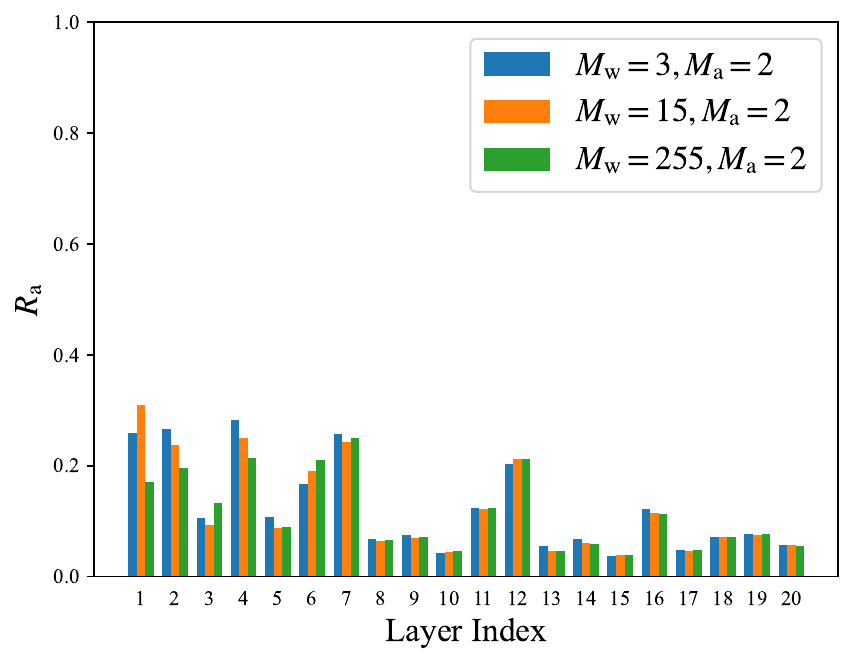}\\
    \caption{Examples of non-zero ratios for each layer when varying $M_{\rm w}$ and fixing $M_{\rm a}$. The dataset is CIFAR-10, and the network architectures are VGG~(top) and PreActResNet~(bottom). Note that since the output layer is not quantized, the graph of $R_{\rm a}$ has one less layer index.}
    \label{fig:results_layer_ma=2}
\end{figure}

\begin{figure}[t]
    \centering
    \includegraphics[width=0.45\linewidth]{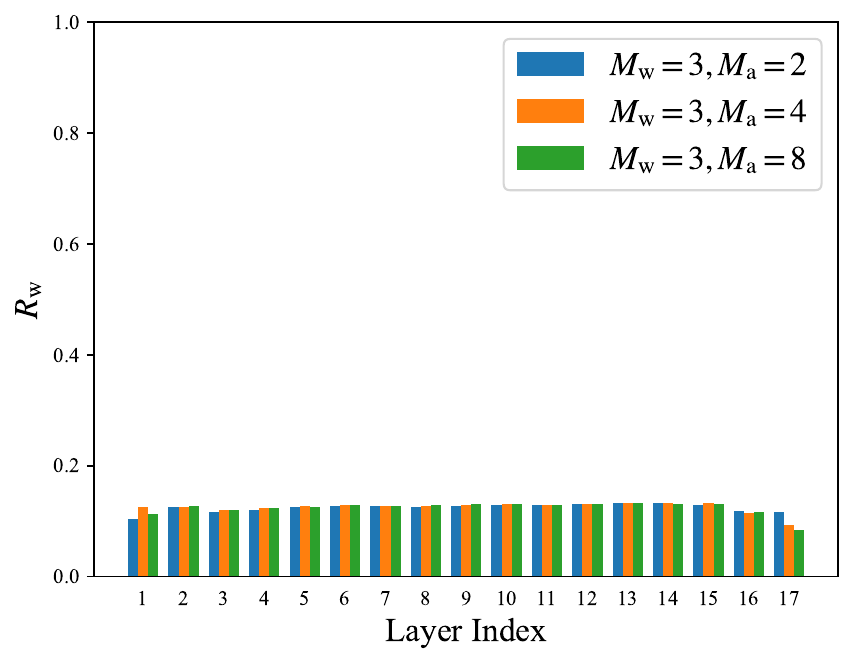}
    \includegraphics[width=0.45\linewidth]{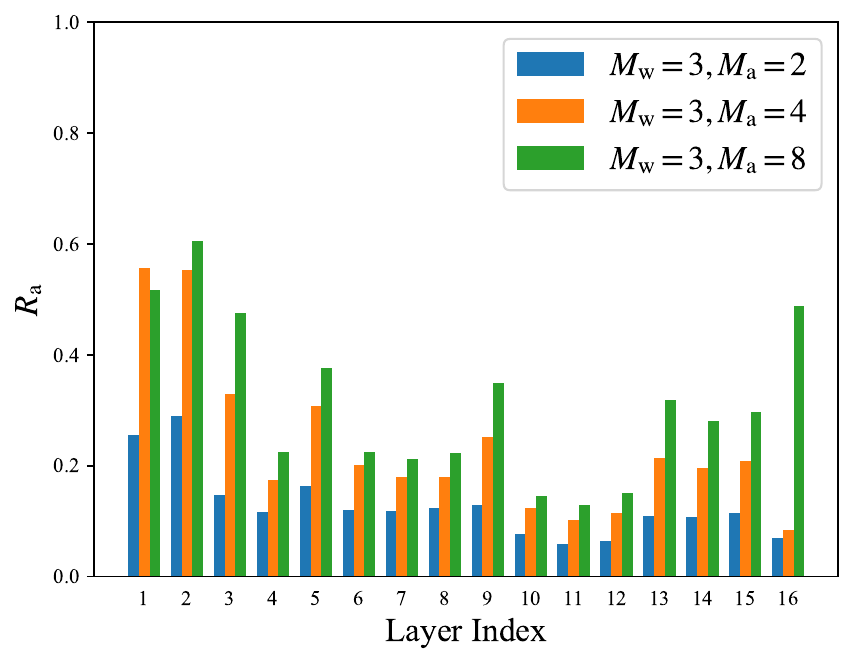}\\
    \includegraphics[width=0.45\linewidth]{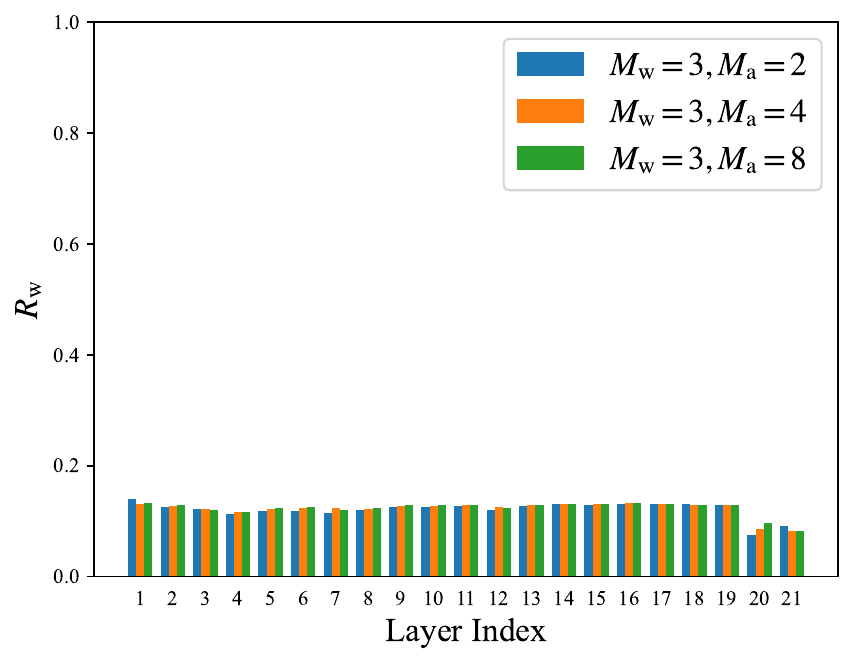}
    \includegraphics[width=0.45\linewidth]{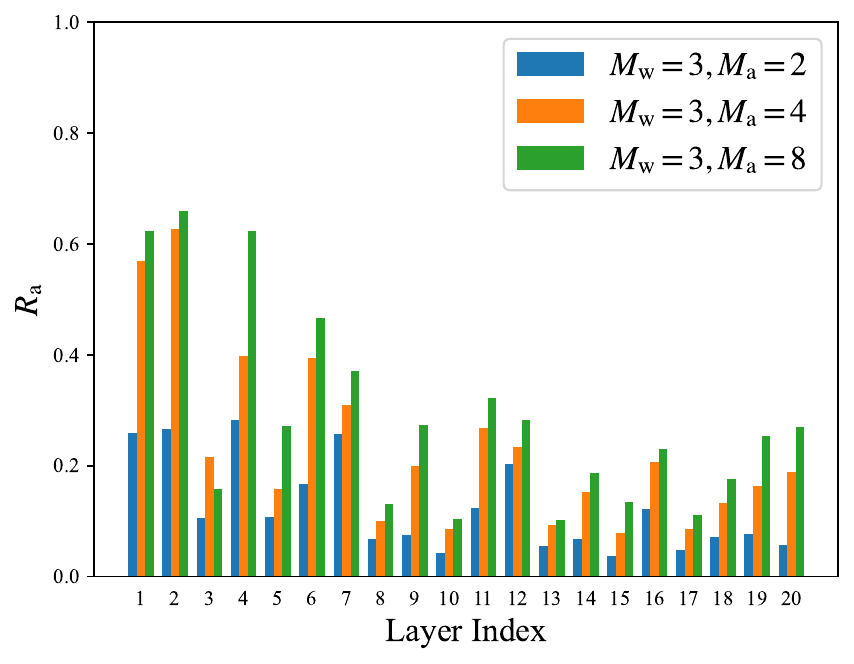}\\
    \caption{Examples of non-zero ratios for each layer when varying $M_{\rm a}$ and fixing $M_{\rm w}$. The dataset is CIFAR-10, and the network architectures are VGG~(top) and PreActResNet~(bottom). Note that since the output layer is not quantized, the graph of $R_{\rm a}$ has one less layer index.}
    \label{fig:results_layer_mw=3}
\end{figure}

Tables~\ref{tbl:vgg-cifar10}, \ref{tbl:resnet-cifar10}, \ref{tbl:vgg-cifar100}, and \ref{tbl:resnet-cifar100} and Figs.~\ref{fig:results_layer_ma=2} and \ref{fig:results_layer_mw=3} show experimental results with respect to changing $M_{\rm w}$ and $M_{\rm a}$.
Here, $R_{\rm w}$ and $R_{\rm a}$ represent the ratio of non-zero parameters of weights and outputs of the activation functions, respectively.
Note that for non-zero activation functions, we compute the value by averaging the non-zero ratio for each test sample.
Also note that when $M_{\rm a} = 2$, the model becomes S$^3$NNs~(Single Step Spiking Neural Networks)~\cite{suetake2023s3nn}.
In this case, the normalization layers can be completely removed during inference by merging them with the weights.
In other words, inference can be done using only additive operations~(details are discussed in~\cite{suetake2023s3nn}).

From these results, the following can be observed:
\begin{itemize}
    \item Setting $M_{\rm w}$ and $M_{\rm a}$ to small values reduces accuracy.
    \item However, sometimes the value relationship is reversed~(e.g., $M_{\rm w} = 255$ and $M_{\rm a} = 8$ v.s. $M_{\rm w} = 15$ and $M_{\rm a} = 8$ in Table~\ref{tbl:vgg-cifar10}). Especially, $M_{\rm w} = 255$ and $M_{\rm a} = \{4, 8\}$ in Table~\ref{tbl:resnet-cifar100} outperforms the non-quantized DNN.
    \item If $M_{\rm w}$ is fixed and $M_{\rm a}$ increases, the value of $R_{\rm w}$ remains almost the same, while the value of $R_a$ increases.
    \item Remarkably, if $M_{\rm a}$ is fixed and $M_{\rm w}$ increases, the value of $R_{\rm w}$ increases, while the value of $R_{\rm a}$ decreases~(especially in layers close to the input).
\end{itemize}

\begin{table}[t]
\caption{Performance with respect to varying $M_{\rm w}$ and $M_{\rm a}$. The dataset is CIFAR-10, and the network architecture is VGG.}
\label{tbl:vgg-cifar10}
\centering
\begin{tabular}{ccccc}
$M_{\rm w}$ & $M_{\rm a}$ & Accuracy~[\%] & $R_{\rm w}$~[\%] & $R_{\rm a}$~[\%] \\
\hline
\multicolumn{2}{c}{Non-quantized} & 95.98 & 100 & 39.38\\
3 & 2 & 92.52 & 12.91 & 14.53\\
3 & 4 & 94.98 & 12.94 & 26.95\\
3 & 8 & 95.32 & 12.95 & 32.97\\
15 & 2 & 94.26 & 82.04 & 14.32\\
15 & 4 & 95.68 & 82.42 & 25.95\\
15 & 8 & 95.93 & 82.36 & 33.41\\
255 & 2 & 94.18 & 99.00 & 13.88\\
255 & 4 & 95.65 & 99.02 & 24.58\\
255 & 8 & 95.89 & 99.02 & 32.31\\
\hline
\end{tabular}
\end{table}

\begin{table}[t]
\caption{Performance with respect to varying $M_{\rm w}$ and $M_{\rm a}$. The dataset is CIFAR-10, and the network architecture is PreActResNet.}
\label{tbl:resnet-cifar10}
\centering
\begin{tabular}{ccccc}
$M_{\rm w}$ & $M_{\rm a}$ & Accuracy~[\%] & $R_{\rm w}$~[\%] & $R_{\rm a}$~[\%] \\
\hline
\multicolumn{2}{c}{Non-quantized} & 95.64 & 100 & 38.02\\
3 & 2 & 93.28 & 12.39 & 13.75\\
3 & 4 & 94.47 & 12.55 & 25.73\\
3 & 8 & 94.98 & 12.69 & 31.91\\
15 & 2 & 94.08 & 82.22 & 13.15\\
15 & 4 & 95.15 & 82.34 & 26.33\\
15 & 8 & 95.23 & 82.20 & 33.99\\
255 & 2 & 93.94 & 99.01 & 11.94\\
255 & 4 & 95.25 & 99.02 & 23.74\\
255 & 8 & 95.30 & 99.01 & 31.25\\
\hline
\end{tabular}
\end{table}

\begin{table}[t]
\caption{Performance with respect to varying $M_{\rm w}$ and $M_{\rm a}$. The dataset is CIFAR-100, and the network architecture is VGG.}
\label{tbl:vgg-cifar100}
\centering
\begin{tabular}{ccccc}
$M_{\rm w}$ & $M_{\rm a}$ & Accuracy~[\%] & $R_{\rm w}$~[\%] & $R_{\rm a}$~[\%] \\
\hline
\multicolumn{2}{c}{Non-quantized} & 76.52 & 100 & 40.89 \\
3 & 2 & 69.19 & 12.51 & 17.35 \\
3 & 4 & 74.02 & 12.64 & 32.45 \\
3 & 8 & 74.85 & 12.71 & 38.54 \\
15 & 2 & 72.27 & 82.78 & 17.10 \\
15 & 4 & 75.44 & 82.58 & 29.23 \\
15 & 8 & 75.66 & 82.49 & 36.33 \\
255 & 2 & 72.73 & 99.04 & 16.75 \\
255 & 4 & 76.08 & 99.03 & 27.51 \\
255 & 8 & 75.86 & 99.03 & 34.47 \\
\hline
\end{tabular}
\end{table}

\begin{table}[t]
\caption{Performance with respect to varying $M_{\rm w}$ and $M_{\rm a}$. The dataset is CIFAR-100, and the network architecture is PreActResNet.}
\label{tbl:resnet-cifar100}
\centering
\begin{tabular}{ccccc}
$M_{\rm w}$ & $M_{\rm a}$ & Accuracy~[\%] & $R_{\rm w}$~[\%] & $R_{\rm a}$~[\%] \\
\hline
\multicolumn{2}{c}{Non-quantized} & 75.47 & 100 & 37.89 \\
3 & 2 & 70.23 & 12.32 & 15.71 \\
3 & 4 & 73.68 & 12.63 & 29.20 \\
3 & 8 & 74.31 & 12.71 & 34.83 \\
15 & 2 & 71.29 & 82.58 & 15.67 \\
15 & 4 & 75.28 & 82.39 & 28.82 \\
15 & 8 & 75.26 & 82.33 & 34.55 \\
255 & 2 & 72.54 & 99.03 & 13.68 \\
255 & 4 & 75.48 & 99.02 & 24.48 \\
255 & 8 & 75.91 & 99.02 & 31.78 \\
\hline
\end{tabular}
\end{table}

These results demonstrate the effectiveness of MaQD, particularly with $M_{\rm w} = 15$ and $M_{\rm a} = 8$, the model can be trained with an accuracy degradation of less than 1\%.
This means that, instead of setting the activation functions to 8 bits as in BitNet b1.58~\cite{ma2024era}, 3 bits may be sufficient.
Also, as mentioned above, when $M_{\rm a}=2$, only additive operations can be used for inference.
In this case, if we can utilize neuromorphic chips like TrueNorth~\cite{akopyan2015truenorth} or Loihi~\cite{davies2018loihi}, we can take advantage of asynchronous processing and significantly reduce power consumption~\cite{suetake2023s3nn}.
That is, the appropriate degrees for quantization should be determined while considering the hardware for inference.

\clearpage
\section{Conclusion}
\label{sec:conclusion}
In this article, we proposed MaQD~(Magic for the Age of Quantized DNNs) based on LBN~(Layer-Batch Normalization), the scaled round-clip function, and the surrogate gradient.
This enables model compression for product integration while reducing machine resource requirements by employing training with small mini-batch sizes.
Applying this method to the image classification tasks, we confirmed its effectiveness.

In the future, the system will be demonstrated in challenging settings, such as the ImageNet dataset~\cite{deng2009imagenet} and LLMs~\cite{brown2020language}, which impose more demanding machine resource requirements, to showcase its versatility. 
It could also be a helpful combination with technologies like GaLore~\cite{zhao2024galore} and Evolutionary Model Merging~\cite{akiba2024evolutionary}, which have the potential to accelerate the democratization of AI development.
Additionally, applications to SNNs~(Spiking Neural Networks)~\cite{ikegawa2022rethinking,maass1997networks} that require time series processing will be explored.

We hope that {\it ubiquitous magic eventually becomes a gift to civilization}.

\bibliographystyle{splncs04}
\bibliography{mybibfile}

\clearpage
\appendix

\section{Network Details}
In this paper, the VGG and PreActResNet used in the quantization experiments were set up as follows.\\\\
VGG: {\small (64C3)$\times$2-(128C3)$\times$2-AP-(256C3)$\times$4-AP-(512C3)$\times$4-AP-(512C3)$\times$4-10C1-GAP},\\
PreActResNet: {\small 64B1-128B1-256B2-256B2-512B2-512B2-512B2-512B2-10C1-GAP},\\\\
where $x$C$y$ represents the convolutional layer with $x$-output channels and $y$-stride, AP represents the average pooling with the kernel size 2, and GAP represents the global average pooling. 
Also, $x$B$y$ represents the residual block, 
\begin{align}
\begin{aligned}
xBy := xSy + xFy, 
\end{aligned}
\end{align}
where
\begin{align}
\begin{aligned}
xSy &:= \text{LBN-ACT-}x\text{C}y\text{-LBN-ACT-}x\text{C}y\text{-LBN},\\
xFy &:= \text{LBN-ACT-}x\text{C}y\text{-LBN},
\end{aligned}
\end{align}
and ACT represents the (quantized) activation function.
Note that the above description of VGG does not show the normalization layer for the simplicity, although the LBN is applied to the features after the weight transformation.
The activation function (not shown above) is then utilized.
Also, the PreActResNet performs AP before processing in blocks where the stride is set to 2  (although this is not explicitly specified too).

\end{document}